\documentclass[journal]{IEEEtran}

\usepackage{cite}

\ifCLASSINFOpdf

\else

\fi

\usepackage{epsfig}
\usepackage{graphicx}
\usepackage{amsmath}
\usepackage{amssymb}
\usepackage{array}
\usepackage{tablefootnote}
\usepackage{subfigure}
\usepackage{float}
\usepackage{caption}
\usepackage{setspace}

\usepackage{algorithm}
\usepackage{algpseudocode}
\usepackage{graphics}

\usepackage{booktabs}
\usepackage{threeparttable}

\usepackage{amsfonts,amssymb} 

\usepackage[square,sort&compress,numbers]{natbib}

\usepackage{amsmath}

\usepackage{epstopdf}

\usepackage[pdftex]{hyperref}

\allowdisplaybreaks

\begin{document}
%
\title{Orthogonal-Coding-Based Feature Generation for Transductive Open-Set Recognition via Dual-Space Consistent Sampling}
%
%
%

\author{Jiayin Sun and Qiulei Dong
\thanks{Manuscript submitted July, 2022. This work is supported by the National Natural Science Foundation of China (U1805264, 61991423, 61972374), and the Strategic Priority Research Program of the Chinese Academy of Sciences (XDB32050100). The corresponding author is Quilei Dong.

Jiayin Sun and Qiulei Dong are with the National Laboratory of Pattern Recognition, Institute of Automation, Chinese Academy of Sciences, Beijing 100190, China, also with the School of Artificial Intelligence, University of Chinese Academy of Sciences, Beijing 100049, China, and also with the Center for Excellence in Brain Science and Intelligence Technology, Chinese Academy of Sciences, Beijing 100190, China (e-mail: jiayin.sun@nlpr.ia.ac.cn; qldong@nlpr.ia.ac.cn).}
}

\markboth{Journal of \LaTeX\ Class Files,~Vol.~XXX, No.~XXX, July~2022}%
{Shell \MakeLowercase{\textit{et al.}}: Bare Demo of IEEEtran.cls for IEEE Journals}

\maketitle

\begin{abstract}
Open-set recognition (OSR) aims to simultaneously detect unknown-class samples and classify known-class samples. Most of the existing OSR methods are inductive methods, which generally suffer from the domain shift problem that the learned model from the known-class domain might be unsuitable for the unknown-class domain. Addressing this problem, inspired by the success of transductive learning for alleviating the domain shift problem in many other visual tasks, we propose an Iterative Transductive OSR framework, called IT-OSR, which implements three explored modules iteratively, including a reliability sampling module, a feature generation module, and a baseline update module. Specifically, at each iteration, a dual-space consistent sampling approach is presented in the explored reliability sampling module for selecting some relatively more reliable ones from the test samples according to their pseudo labels assigned by a baseline method, which could be an arbitrary inductive OSR method. Then, a conditional dual-adversarial generative network under an orthogonal coding condition is designed in the feature generation module to generate discriminative sample features of both known and unknown classes according to the selected test samples with their pseudo labels. Finally, the baseline method is updated for sample re-prediction in the baseline update module by jointly utilizing the generated features, the selected test samples with pseudo labels, and the training samples. Extensive experimental results on both the standard-dataset and the cross-dataset settings demonstrate that the derived transductive methods, by introducing two typical inductive OSR methods into the proposed IT-OSR framework, achieve better performances than 15 state-of-the-art methods in most cases.
\end{abstract}


%

\section{Introduction}    \label{section:introduction}
%
%
%
%

\IEEEPARstart{I}{n} many real-world scenarios, the deployment of image recognition models is required under open-set conditions. This encourages more and more researchers in the fields of pattern recognition and computer vision to investigate the open-set recognition (OSR) problem, which aims to correctly identify unknown-class samples and maintain the classification accuracy for known-class samples~\cite{openmax, DOC1, Dis1, CAC, Placeholder, CROSR, C2AE, GDFR, Reciprocal, A-RPL, extreme_geometric, S2OSC, CGDL, Hybrid, G-openmax, OSRCI, OpenGAN, Capsule}. 

Existing OSR methods can be roughly divided into two groups: the discriminative OSR methods~\cite{openmax, DOC1, Dis1, CAC, Placeholder, CROSR, C2AE, GDFR, Reciprocal, A-RPL, extreme_geometric, S2OSC} which directly learn the classifiers for recognizing different classes based on the discriminative representations, and the generative OSR methods~\cite{CGDL, Hybrid, G-openmax, OSRCI, Capsule} which model the distributions of the known-class samples. However, most of these methods~\cite{openmax, DOC1, Dis1, CAC, Placeholder, CROSR, C2AE, GDFR, Reciprocal, A-RPL, extreme_geometric, CGDL, Hybrid, G-openmax, OSRCI, Capsule} employ the inductive OSR strategy, which learns recognition models from a training dataset consisting of labeled known-class samples. And they have to suffer from the domain shift problem that their learned recognition models from the known-class domain are not suitable for the unknown-class domain. 

Different from the inductive learning strategy, the transductive learning strategy, which makes use of both training and test samples for model learning, has demonstrated its effectiveness for alleviating the domain shift problem recently in various visual tasks~\cite{ZSL_transductive1, ZSL_transductive2, segtrans1, segtrans2, tracktrans1}. But to our knowledge, only a pioneering work~\cite{S2OSC} investigated transductive OSR, where a model (called S2OSC) was trained in a transductive manner. S2OSC~\cite{S2OSC} firstly provided pseudo labels for the test samples that were predicted as unknown classes with high confidence scores by a baseline method, and then the baseline model was re-trained by jointly utilizing the pseudo-labeled test samples, the remaining unlabeled test samples, and the labeled known-class samples from the original training set. However, the following two open problems still remain for the transductive OSR task:
\begin{itemize}
	\item[Q1:] The pseudo-labeled data selection problem: In general, some test samples are inevitably misclassified during the transductive process at the training stage, resulting in a poor classification model by utilizing them for training. This issue raises the problem: How to select a relatively reliable subset from the whole set of the pseudo-labeled test samples for model training?
	\item[Q2:] The sample imbalance problem: The number of the pseudo-labeled unknown-class test samples at the training stage is generally much smaller than that of the known-class training samples, so that it is prone to train a poor classification model by jointly utilizing these imbalanced samples. This issue raises the problem: How to automatically obtain a larger and balanced number of data from the given imbalanced samples for model training?
\end{itemize}

Addressing the aforementioned domain shift problem as well as Problems Q1 and Q2, we propose an Iterative Transductive OSR framework, called IT-OSR, where three explored modules (a reliability sampling module, a feature generation module, and a baseline update module) are implemented in an iterative manner. At each iteration, addressing Problem Q1, a dual-space consistent sampling approach, which takes both the score output space and a latent feature space into account, is proposed for selecting a relatively more reliable subset of test samples from the test dataset in the reliability sampling module. Then, a conditional dual-adversarial generative network is designed to generate an arbitrary number of both known-class and unknown-class features for alleviating the sample imbalance problem (\emph{i.e.} Problem Q2) in the feature generation module. Finally, the baseline method is updated and the labels of the test samples are re-predicted for the next iteration in the baseline update module.

In sum, the main contributions of this paper are three-fold: 

\begin{enumerate}
	\item[-] We explore the dual-space consistent sampling approach for sample selection, which could select a relatively reliable subset from all the test samples for handling the aforementioned Problem Q1.
	
	\item[-] We design the conditional dual-adversarial generative network under an orthogonal coding condition. It could generate an arbitrary number of known-class and unknown-class sample features, guaranteeing that the numbers of both the generated known-class and unknown-class sample features are sufficiently large and balanced for alleviating the aforementioned Problem Q2.	
	
	\item[-] We propose the IT-OSR framework, based on the explored dual-space consistent sampling approach and the designed conditional dual-adversarial generative network. The proposed IT-OSR framework could accommodate an arbitrary inductive OSR method as its baseline method, and its effectiveness is demonstrated by the experimental results in Section~\ref{Experiments}.
	
\end{enumerate}

The remainder of this paper is organized as follows. Some existing inductive OSR methods are reviewed in Section~\ref{Related_Work}. The proposed framework is described in detail in Section~\ref{Method}. Experimental results are reported in Section~\ref{Experiments}. Section~\ref{Conclusion} concludes the paper.

\section{Related Works}    \label{Related_Work}
As discussed above, to our knowledge, only a pioneering transductive OSR method~\cite{S2OSC} has been proposed in literature. Considering that a transductive method generally needs to use an inductive method as its baseline method, hence, we give a detailed review on the existing works for inductive OSR from the following two aspects.

\textbf{Discriminative inductive OSR Methods.} The discriminative methods directly learn more generalized classifiers through elaborately designed losses or modules, then they classify the known-class samples and detect the unknown-class samples by their discriminative network representations. Bendale and Boult~\cite{openmax} proposed OpenMax, and they modified the traditional closed-set classification network by replacing the SoftMax layer with their proposed OpenMax layer which calculated open-set probabilities based on the distances to each known-class center. Hassen and Chan~\cite{Dis1} utilized Linear Discriminant Analysis for learning more compact known-class representations such that more space could be reserved for the unknown classes for alleviating the confusion between known classes and unknown classes. Similarly, Miller \emph{et al.}~\cite{CAC} proposed class anchor clustering loss, which constrained the known-class representations in the logit space. Zhou \emph{et al.}~\cite{Placeholder} proposed to reserve placeholders for unknown classes during model training by adding an additional output and mixuping features. Yoshihashi \emph{et al.}~\cite{CROSR} proposed a joint training framework where classification and reconstruction were simultaneously implemented, and the model prediction vector concatenated with latent features was used for classification. Oza and Patel~\cite{C2AE} proposed a class conditioned autoencoder, which detected unknown classes by training with reconstruction errors of the mismatched-class image pairs. Perera \emph{et al.}~\cite{GDFR} proposed to add the reconstructed images of the known classes as additional channels of the 3-channel input images for model training. Chen \emph{et al.}~\cite{Reciprocal} proposed to learn discriminative reciprocal points by digging 1-vs-rest information among the known classes for reserving the extra-class space as the open space, and they also proposed an adversarial version called A-RPL in~\cite{A-RPL}. Perera and Patel~\cite{extreme_geometric}
proposed to train images ensembled with different transformed versions by extreme geometric transformations for mining more information from the original known-class images.

\textbf{Generative inductive OSR Methods.} The generative methods utilize either GANs~\cite{GAN} or other generative models~\cite{VAE, resflow} to model the distributions of the known classes. Some methods modeled the known-class distributions either explicitly or implicitly, and then detected the unknown classes by their inconsistency with these distributions. Sun \emph{et al.}~\cite{CGDL} proposed to model each known class to be subject to a Gaussian, thus the representations of the test samples deviating from the modeled Gaussians could be regarded as unknown classes. Similarly, Guo \emph{et al.}~\cite{Capsule} proposed to use a capsule network to better model each known class as a Gaussian. Zhang \emph{et al.}~\cite{Hybrid} proposed to apply a flow-based model for hybrid training of both classification and generation, whose model could output the probability of the test sample belonging to the known classes. Kong and Ramanan~\cite{OpenGAN} utilized the discriminator trained by the adversarial game between the known-class images/features and the outlier images/features after model selection, the output of which is utilized for detecting unknown classes. Besides, some methods further generated unknown-class samples based on the modeled known-class distributions, and then trained open-set classifiers using these generated unknown-class samples as an additional class. Neal \emph{et al.}~\cite{OSRCI} proposed to generate counterfactual images as unknown classes which were easily confused with the known classes in the feature space. Chen \emph{et al.}~\cite{A-RPL} also proposed an enhanced version from A-RPL, called ARPL-CS, which generated diverse and confusing samples for data augmentation.

\begin{figure*}[t]
	\begin{center}
		\setlength{\abovecaptionskip}{0.cm}
		\includegraphics[height=6cm,width=15.6cm]{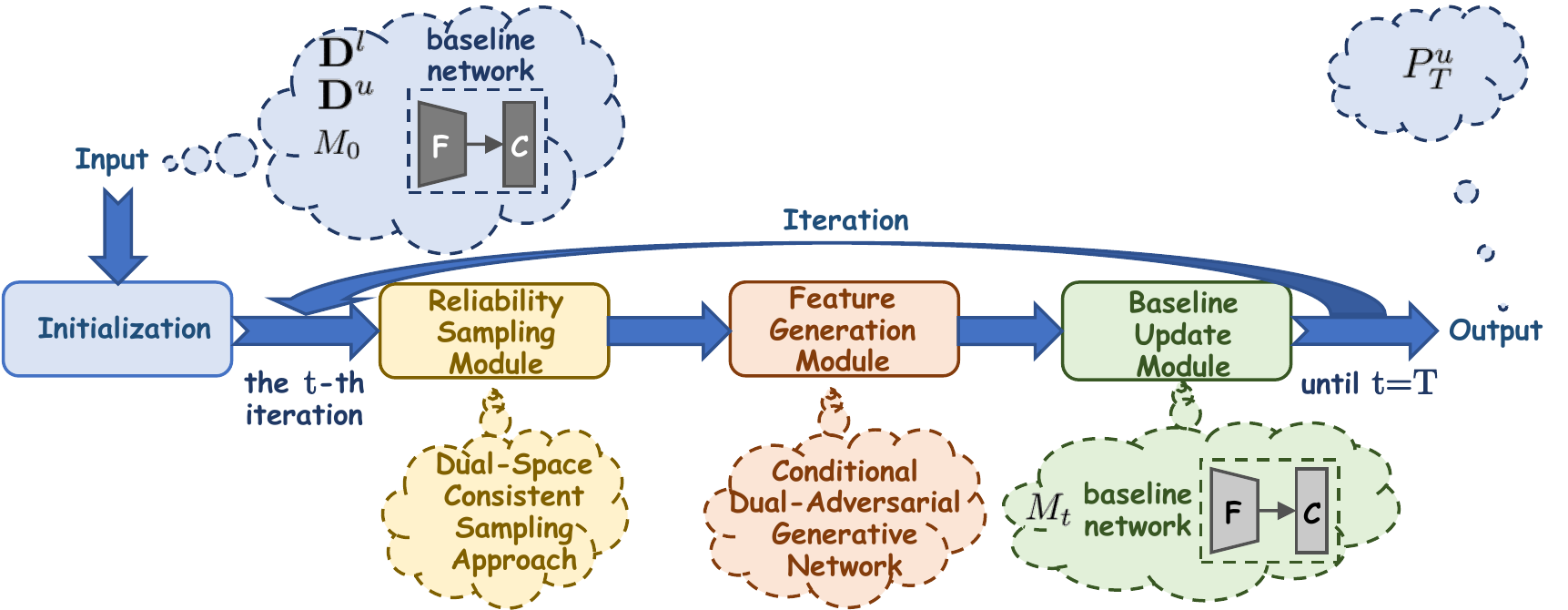}
	\end{center}
	\caption{Pipeline of the proposed IT-OSR framework which implements the reliability sampling module, the feature generation module, and the baseline update module iteratively.}
	\label{fig:overall_scheme}
\end{figure*}

\section{Methodology}  \label{Method}
In this section, we propose the IT-OSR framework that iteratively implements three explored modules: a reliability sampling module, a feature generation module, and a baseline update module. Firstly, the pipeline of the proposed IT-OSR is introduced. Then, the three explored modules are described respectively in detail. Finally, two novel transductive OSR methods are derived from the proposed IT-OSR framework.

\subsection{Pipeline of IT-OSR}
The pipeline of the proposed IT-OSR framework, consisting of a reliability sampling module, a feature generation module, and a baseline update module, is shown in Figure~\ref{fig:overall_scheme}. As seen from this figure, given an arbitrary inductive OSR method as the baseline method whose network $M_0$ can always be divided into two parts: a feature extractor $F$ and a linear/nonlinear classifier $C$, all the test samples are initially classified such that the features outputted from $F$ and the pseudo labels of the test samples are obtained by this baseline at the initialization stage. 

Then, the three explored modules are implemented in an iterative manner: At the $t$-th ($t=1,2...,T$, and $T$ represents a preseted maximum iteration number) iteration, firstly in the reliability sampling module, a relatively reliable subset $ \mathbf{D}_{s_t}^p $ of the test samples is selected from the unlabeled test dataset $ \mathbf{D}^u $ by an explored dual-space consistent sampling approach. Then in the feature generation module, a synthesized set $ \mathbf{D}_{g_t} $ containing both known-class and unknown-class sample features is generated by a designed conditional dual-adversarial generative network which is trained on the union set $ \mathbf{D}^l \cup \mathbf{D}_{s_t}^p $ of the training set and the sampled test set to balance the numbers of the training samples of between known and unknown classes. Finally in the baseline update module, the baseline model is updated (re-trained) according to the union set $ \mathbf{D}^l \cup \mathbf{D}_{s_t}^p \cup \mathbf{D}_{g_t} $ for making predictions $P_t^u$ on the test dataset $ \mathbf{D}^u $. The iterative process would not be terminated until the iteration number $t$ reaches a preseted maximum $T$, and the assigned labels to all the test samples by the re-trained baseline method at the final iteration are used as the final predictions. The whole above process of IT-OSR is also outlined in Algorithm~\ref{code:overall_framework}. In the following parts, the three explored modules at each iteration in Figure~\ref{fig:overall_scheme} (\emph{i.e.} the three key Steps 3-5 in Algorithm~\ref{code:overall_framework}) are introduced respectively.

\begin{algorithm}[t]
	\caption{The IT-OSR framework} 
	\renewcommand{\algorithmicrequire}{\textbf{Input:}}
	\renewcommand{\algorithmicensure}{\textbf{Output:}}
	\begin{algorithmic}[1] 
		\Require The labeled training set $ \mathbf{D}^l $, the unlabeled test dataset $ \mathbf{D}^u $, and the initial baseline model $M_0$
		\Ensure The predictions $P_T^u$ on the test dataset $ \mathbf{D}^u $ maded by the updated model at the $T$-th iteration
		\State Initialization: Make predictions on $ \mathbf{D}^u $ by $M_0$;
		\For{$t=1$ to $T$} 
		\State \textit{Reliability Sampling Module}: Select a subset $\mathbf{D}_{s_t}^p$ of test samples from $ \mathbf{D}^u $ based on the dual-space consistent sampling approach;
		\State \textit{Feature Generation Module}: Train the conditional dual-adversarial generative network under an orthogonal coding condition with $ \mathbf{D}^l \cup \mathbf{D}_{s_t}^p $ and generate the generated set $ \mathbf{D}_{g_t} $ of features from the generator;
		\State \textit{Baseline Update Module}: Obtain the updated baseline model $M_t$ by re-training the baseline model with $ \mathbf{D}^l \cup \mathbf{D}_{s_t}^p \cup \mathbf{D}_{g_t} $ and make predictions $P_t^u$ on $ \mathbf{D}^u $ by $M_t$;
		\EndFor 
		\State \Return $P_T^u$;
	\end{algorithmic} \label{code:overall_framework}
\end{algorithm}

\begin{figure*}[t]
	\begin{center}
		\setlength{\abovecaptionskip}{0.cm}
		\includegraphics[height=6cm,width=14.4cm]{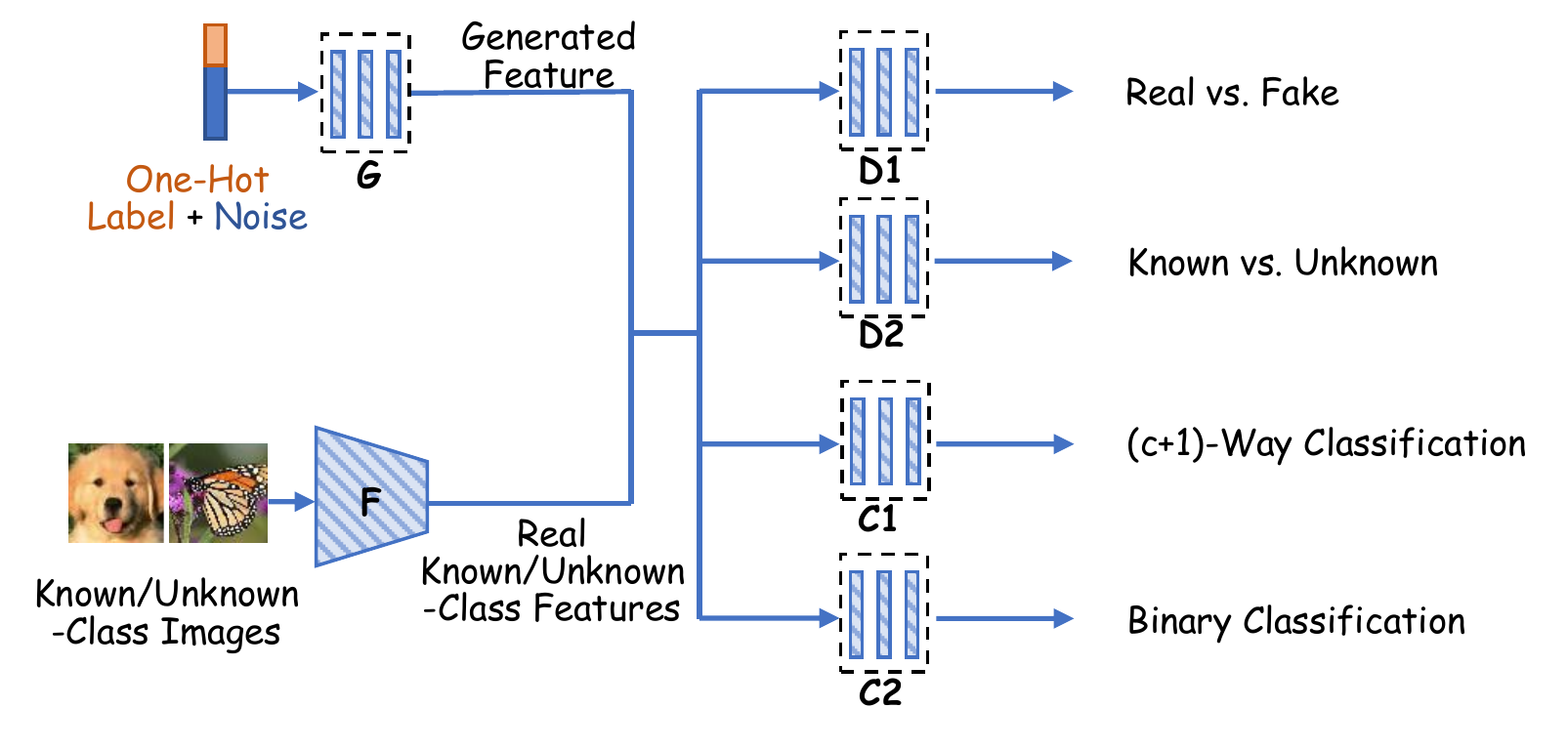}
	\end{center}
	\caption{The architecture of the conditional dual-adversarial generative network under an orthogonal coding condition.}
	\label{fig:improved_cGAN}
\end{figure*}

\subsection{Reliability Sampling Module}
In this module, addressing the referred pseudo-labeled data selection problem in Section~\ref{section:introduction}, a dual-space consistent sampling approach is explored for selecting a relatively reliable subset from the test dataset, given the predictions of the baseline model at the previous iteration. Here, considering that some of the test samples are inevitably misclassified by the constantly updated baseline network at each iteration, only such test samples that have consistent predictions between the following two defined spaces, are considered as relatively `reliable' samples by the explored dual-space consistent sampling approach according to the following definitions and criterion:

\noindent\textbf{Definition 1 (Score Output Space).} \textit{The score output space is defined as the space spanned by the score vectors outputted from the classifier $C$ of a baseline network.}

\noindent\textbf{Definition 2 (Latent Feature Space).} \textit{The latent feature space is defined as the space 
spanned by the features outputted from the feature extractor $F$ of a baseline network.}

\noindent\textbf{Criterion.} \textit{A test sample, whose pseudo label assigned in the score output space is consistent with those of more than half of its spatial neighbors in the latent feature space, is identified as a relatively reliable sample.}

According to this criterion, the dual-space consistent sampling approach is implemented in the following two steps:


\begin{itemize}
	
	\item [S1:] Given the pseudo labels and confidence scores $s$ calculated in the score output space by the updated baseline model at the previous iteration, the test samples are classified into three groups: \textit{known class}, \textit{unknown class}, and \textit{undetermined class} according to the pre-defined threshold which depends on the statistical values calculated from the confidence scores of the training set $\mathbf{D}^l$, and the three groups are denoted as $\mathbf{D}_{k}^{p}$, $\mathbf{D}_{unk}^{p}$, and $\mathbf{D}_{und}^{p}$ respectively. The above grouping process can be formulated as:
	\begin{align}
	\left\{
	\begin{aligned}
	&\mathbf{D}_{k}^{p}, \ \ \; \, \, \mathrm{if} \ s > \mu+\alpha  \delta    \\
	&\mathbf{D}_{unk}^{p}, \ \mathrm{if} \ s < \mu-\alpha \delta   \\
	&\mathbf{D}_{und}^{p}, \ \mathrm{if} \ \mu-\alpha \delta \leq s  \leq \mu+\alpha \delta  
	\end{aligned}
	\right. \label{threshold}
	\end{align}
	where $\mu$ and $\delta$ are the mean and standard deviation of the confidence scores calculated from the training set $\mathbf{D}^l$ respectively, and $\alpha$ is a hyper-parameter.

	\item [S2:] For each sample in $\mathbf{D}_{k}^{p} \cup \mathbf{D}_{unk}^{p}$, we search for its $K$ nearest neighbors from $\mathbf{D}_{k}^{p} \cup \mathbf{D}_{unk}^{p} \cup \mathbf{D}_{und}^{p}$ in the latent feature space according to the Euclidean distances. Then, the test samples from $\mathbf{D}_{k}^{p} \cup \mathbf{D}_{unk}^{p}$ whose pseudo labels are consistent with that of more than their $K/2$ neighbors are retained, while those inconsistent with more than their $K/2$ neighbors are removed out of $\mathbf{D}_{k}^{p} \cup \mathbf{D}_{unk}^{p}$. This selected subset of test samples is denoted as $\mathbf{D}_{s_t}^p$.
	
\end{itemize}


\subsection{Feature Generation Module} 
In this module, we design a conditional dual-adversarial generative network under an orthogonal coding condition for generating discriminative known-class and unknown-class sample features, based on the features of both the training samples and the selected test samples outputted from the feature extractor $F$ of the updated baseline model. The designed network simply utilizes one-hot vector as the orthogonal coding condition, and introduces this condition to the generative adversarial network in a similar manner to ACGAN~\cite{ACGAN}. It contains a generator $G$, a feature extractor $F$, two discriminators $D_1$ and $D_2$, two classifiers $C_1$ and $C_2$. The architecture of the conditional dual-adversarial generative network is shown in Figure~\ref{fig:improved_cGAN}.

\textbf{Generator $G$:} As seen from Figure~\ref{fig:improved_cGAN}, the generator $G$ is designed to generate known/unknown-class sample features $ \mathbf{D}_{g_t} $, whose input is the orthogonal coding condition concatenated with a Gaussian noise vector. Here, the generator $G$ has a three-layer perceptron architecture. It contains 3 fully-connected layers and uses ReLU (Rectified Linear Unit)~\cite{ReLU} as the nonlinear activation function, and the dimensionality of its hidden layers is 4096.

\textbf{Feature extractor $F$:} The feature extractor $F$ is straightforwardly from the updated baseline model at the previous iteration, whose weights are fixed during the training process of the generative network.

\textbf{Discriminator $D_1$:} The discriminator $D_1$ is designed to implement true or false discrimination, whose inputs are the fake features generated by $G$ and the real image features outputted by the feature extractor $F$, and the dimensionality of its outputs is 1. Here, the discriminator $D_1$ has the same three-layer perceptron architecture as $G$ except the dimensionalities of the inputs and the outputs.

\textbf{Discriminator $D_2$:} The discriminator $D_2$ is designed to implement known-class or unknown-class discrimination for further improving the discriminating ability of the generated features, whose inputs contain known/unknown-class real/generated features, and the dimensionality of its outputs is 1. Here, the discriminator $D_2$ has the same three-layer perceptron architecture as $D_1$.


\textbf{Classifier $C_1$:} The classifier $C_1$ is designed to classify both the real and the generated features into $(c+1)$ classes ($c$ known classes and one \emph{unknown class}). Similar to ACGAN~\cite{ACGAN}, this classifier is used to strengthen the each-class distinguishability of both the real and the generated features. Here, the classifier $C_1$ has the same three-layer perceptron architecture as $D_1$ except the dimensionality of the outputs. The dimensionality of its outputs is $(c+1)$.

\textbf{Classifier $C_2$:} The classifier $C_2$ is designed to identify whether an input real/generated feature belongs to a known class or not. This is to say, it classifies both the real and generated features into two groups (\emph{known class} and \emph{unknown class}). This classifier is used to pay more attention to the distinguishability of the unknown-class features from the known-class features. Here, the classifier $C_2$ has the same three-layer perceptron architecture as $C_1$ except the dimensionality of the outputs. The dimensionality of its outputs is 2.

Similar to WGAN~\cite{WGAN}, the adversarial game between the generator $G$ and the discriminator $D_1$ is defined as:
\begin{align}
&\min_{G}\max_{D_1}V(G, D_1) = 
\mathbb{E}_{x \sim P_r} \left[ D_1(x) \right]  -\mathbb{E}_{\tilde{x} \sim P_g} \left[ D_1(\tilde{x}) \right]
\end{align}
where $\mathbb{E}$ is the expectation, $P_r$ and $P_g$ are the data distribution and the generator's distribution respectively, $x$ and $\tilde{x}$ are the extracted real image features and the generated fake features respectively.

Similar to ACGAN~\cite{ACGAN}, the classification loss for the classifier $C_1$ is defined as:
\begin{align}
\begin{aligned}
&\mathcal{L}_{C_1} = - \mathbb{E}_{x \sim P_r} \left[ \mathrm{log}P (y_{c_1}=y_{l_1}|x) \right] \\
&\phantom{=\ \; \; \; \; \, \, } - \mathbb{E}_{\tilde{x} \sim P_g} \left[ \mathrm{log}P (y_{c_1}=y_{l_1}|\tilde{x}) \right]   \label{12}
\end{aligned}
\end{align}
where $y_{c_1}$ are the predicted labels predicted by $C_1$, $y_{l_1}$ are the ground truths for $(c+1)$-way classification.

In addition, the adversarial game between the generator $G$ and the discriminator $D_2$ is defined as:
\begin{align}
\begin{aligned}
&\min_{G}\max_{D_2}V(G, D_2) = \\	
&- \mathbb{E}_{x \sim P_r} \left[ D_2(x|y=c+1) - D_2(x|y \neq c+1) \right] \\
& - \mathbb{E}_{\tilde{x} \sim P_g} \left[ D_2(\tilde{x}|\tilde{y}=c+1) - D_2(\tilde{x}|\tilde{y} \neq c+1) \right]
\end{aligned}
\end{align}
where $c$ is the number of the known classes, $y$ and $\tilde{y}$ are the class conditions of $x$ and $\tilde{x}$ respectively. 

The binary classification loss for the classifier $C_2$ is defined as:
\begin{align}
\begin{aligned}
&\mathcal{L}_{C_2} = - \mathbb{E}_{x \sim P_r} \left[ \mathrm{log}P (y_{c_2}=y_{l_2}|x) \right] \\
&\phantom{=\ \; \; \; \; \, \, } - \mathbb{E}_{\tilde{x} \sim P_g} \left[ \mathrm{log}P (y_{c_2}=y_{l_2}|\tilde{x}) \right]   \label{11}
\end{aligned}
\end{align}
where $y_{c_2}$ are the predicted labels predicted by $C_2$, and $y_{l_2}$ are the ground truths for binary classification.

Accordingly, the dual-adversarial game is to optimize the objective function $V(G, D_1, D_2)$, which is the weighted sum of the aforementioned two adversarial objective functions $V(G, D_1)$ and $V(G, D_2)$:
\begin{align}
\begin{aligned}
&\min_{G}\max_{D_1}\max_{D_2}V(G, D_1, D_2) = \\
&\phantom{=\, } \mathbb{E}_{x \sim P_r} \left[ D_1(x) \right]  -\mathbb{E}_{\tilde{x} \sim P_g} \left[ D_1(\tilde{x}) \right] +  \\
&\phantom{=\, } \lambda \left\{ - \mathbb{E}_{x \sim P_r} \left[ D_2(x|y=c+1) - D_2(x|y \neq c+1) \right]  \right.  \\
&\phantom{=\ \ \; \, } \left. - \mathbb{E}_{\tilde{x} \sim P_g} \left[ D_2(\tilde{x}|\tilde{y}=c+1) - D_2(\tilde{x}|\tilde{y} \neq c+1) \right] \right\}    \label{8}
\end{aligned}
\end{align}
where $\lambda$ is a hyper-parameter balancing the weights of the two adversarial games.

Besides, the total classification loss is the sum of the aforementioned two classification losses:
\begin{align}
\begin{aligned}
&\mathcal{L}_{cls} = \mathcal{L}_{C_1} + \mathcal{L}_{C_2}
\end{aligned}
\end{align}

\subsection{Baseline Update Module}
In this module, given the training samples, the generated features, and the selected test samples with pseudo labels at the $t$-th iteration, the baseline method is re-trained by utilizing the real/generated unknown-class features as an additional class (the $(c+1)$-th class) for classification, and then it predicts the pseudo labels $ P_t^u $ of all the test samples for the next iteration.

%
%
%

\subsection{Two Transductive OSR Methods Derived From IT-OSR}
It is noted that the proposed IT-OSR framework could accommodate not only an arbitrary existing inductive OSR method but also a new inductive one, resulting in a transductive OSR method. For evaluating the effectiveness of the proposed IT-OSR framework in the following section, we explore the following two transductive methods:

\begin{itemize}
	
	\item [1)] \textbf{IT-OSR-ARPL.} This transductive method is derived from the IT-OSR framework by simply utilizing the existing inductive method A-RPL~\cite{A-RPL} as the baseline method.
	
	
	\item [2)] \textbf{IT-OSR-TransP}. We firstly design a new inductive OSR network by simply concatenating the Swin Transformer~\cite{Swin_Transformer} (used as the feature extractor $F$) and a three-layer perceptron (used as the classifier $C$ that has the similar architecture to the aforementioned classifier $C_1$), called TransP, which is trained with the traditional cross-entropy classification loss. Then, the transductive method IT-OSR-TransP is derived from the proposed IT-OSR framework by utilizing the designed TransP as the baseline method.
	
	
\end{itemize}

\section{Experiments}  \label{Experiments}
In this section, firstly, we give a brief introduction on the standard-dataset setting, the cross-dataset setting, and their evaluation metrics. Then, we show the implementation details in our experiments. Next, we evaluate the proposed IT-OSR framework. Finally, we conduct the ablation study.

\subsection{Dataset Settings and Evaluation Metrics}

\subsubsection{Dataset Settings}
In order to assess open-set recognition performance under different degrees of domain shift, we conduct experiments under two dataset settings: the standard-dataset setting where both the known-class and unknown-class samples are obtained from a same dataset, and the cross-dataset setting where the known-class and unknown-class samples are obtained from two different datasets respectively.

Under the standard-dataset setting, the following six standard datasets are used for evaluation as done in~\cite{openmax, CAC, Placeholder, CROSR, C2AE, GDFR, Reciprocal, extreme_geometric, CGDL, Hybrid, OSRCI, A-RPL, Capsule, S2OSC}:

\begin{itemize}
	
	\item [-] 
	\textbf{MNIST~\cite{MNIST}:} MNIST is a traditional benchmark dataset for classification containing 10 categories of handwritten digit images, in which 6 categories are randomly chosen as the known classes, while the rest 4 categories as the unknown classes.
	
	\item [-]
	\textbf{SVHN~\cite{SVHN}:} Similar to MNIST, SVHN also contains 10 categories of digit images but from street view house numbers. It is likewise divided into 6 known classes and 4 unknown classes.
	
	\item [-]
	\textbf{CIFAR10~\cite{CIFAR10}:} CIFAR10 is made up of 10 categories of natural images, 6 of which act as the known classes, while the rest 4 act as the unknown classes.
	
	\item [-]
	\textbf{CIFAR+10/+50:} CIFAR+10/+50 contains two datasets based on different combinations of CIFAR10 and CIFAR100~\cite{CIFAR100}. They both randomly select 4 vehicle categories in CIFAR10 as the known classes, and 10/50 random categories in CIFAR100 act as the unknown classes in CIFAR+10/+50.
	
	\item [-]
	\textbf{TinyImageNet~\cite{TinyImageNet}:} TinyImageNet is a more complex dataset that contains 200 categories of ImageNet~\cite{ImageNet}, 20 of which are randomly chosen as the known classes, while the rest 180 categories act as the unknown classes.

\end{itemize}
For a fair comparison, we use the same data splits as done in~\cite{OSRCI, GDFR, Hybrid, extreme_geometric}.

Under the cross-dataset setting, the whole 10 categories in the CIFAR10 dataset act as the known classes, while two datasets, TinyImageNet and LSUN~\cite{LSUN}, are either cropped or resized for acting as the unknown classes respectively, as done in~\cite{openmax, Placeholder, CROSR, C2AE, GDFR, CGDL, OSRCI, Capsule}.

\subsubsection{Evaluation Metrics}
Under the standard-dataset setting, the AUROC and ACC are utilized as the evaluation metrics for evaluating the performance on detecting unknown classes and classifying known classes respectively as done in~\cite{openmax, CAC, Placeholder, CROSR, C2AE, GDFR, Reciprocal, extreme_geometric, CGDL, OSRCI, A-RPL, Capsule}:

\begin{itemize}
	
	\item [-] 
	\textbf{AUROC:} The Receiver Operating Characteristic (ROC) Curve is depicted by the False Positive Rate (FPR) as abscissa and the True Positive Rate (TPR) as vertical coordinate. And the area under ROC curve (AUROC) is a typical evaluation metric for open-set detection, which is not affected by the threshold chosen for separating between the two classes. 

	\item [-]
	\textbf{ACC:} The top-1 accuracy (ACC) is a typical evaluation metric in closed-set classification.
	
\end{itemize}

Under the cross-dataset setting, the macro-F1 score is utilized as the evaluation metric for evaluating the performance on classifying both known classes and unknown classes simultaneously as done in~\cite{openmax, Placeholder, CROSR, C2AE, GDFR, CGDL, OSRCI, Capsule}:
\begin{itemize}
	
	\item [-] 
	\textbf{macro-F1 score:} The macro-F1 score measures the ($c+1$)-way classification performance, which is not influenced by data imbalance. 
	
\end{itemize}

\subsection{Implementation Details}
In all of our experiments, we use the feature extractor in Swin-B~\cite{Swin_Transformer} pre-trained by ImageNet-22K~\cite{ImageNet} at $t=1$, and use the feature extractor of $M_t$ at $t>1$ for the latent feature space. The $\alpha$ in Equation~(\ref{threshold}) is set to $2.5$, the number of nearest neighbors in Step S2 in the reliability sampling module is set to $K=10$, the balancing weight $\lambda$ in Equation~(\ref{8}) is set to $0.1$, and the maximum iteration number $T$ is set to $T=2$. In training, we use the SGD optimizer with the learning rate $0.002$ for updating the feature extractor $F$ of the baseline network, and $0.02$ for updating the classifier part of the baseline network and training the generative network.

\begin{table*}
	\centering
	\setlength{\abovecaptionskip}{0pt}
	\setlength{\belowcaptionskip}{10pt}
	\caption{Evaluation on open-set detection (AUROC) under the standard-dataset setting. The reported results are averaged over the same five trials as~\cite{OSRCI, GDFR, Hybrid, extreme_geometric}.}
	\begin{tabular}{m{2.4cm}<{\raggedright}m{1.2cm}<{\centering}m{1.2cm}<{\centering}m{1.5cm}<{\centering}m{1.5cm}<{\centering}m{1.8cm}<{\centering}m{2.2cm}<{\centering}m{2.2cm}<{\centering}}
		\toprule
		Method & Inductive & Transductive & MNIST & SVHN & CIFAR10 & CIFAR+10/+50 & TinyImageNet  \\
		\midrule
		SoftMax & $\surd$ & $\times$ & 0.978 & 0.886 & 0.677 & 0.816/0.805 & 0.577  \\
		OpenMax~\cite{openmax} & $\surd$ & $\times$ & 0.981 & 0.894 & 0.695 & 0.817/0.796 & 0.576  \\
		OSRCI~\cite{OSRCI} & $\surd$ & $\times$ & 0.988 & 0.910 & 0.699 & 0.838/0.827 & 0.586  \\
		CROSR~\cite{CROSR} & $\surd$ & $\times$ & 0.991 & 0.899 & 0.883 & 0.912/0.905 & 0.589  \\
		C2AE~\cite{C2AE} & $\surd$ & $\times$ & 0.989 & 0.922 & 0.895 & 0.955/0.937 & 0.748  \\
		CGDL~\cite{CGDL} & $\surd$ & $\times$ & 0.994 & 0.935 & 0.903 & 0.959/0.950 & 0.762  \\
		GDFR~\cite{GDFR} & $\surd$ & $\times$ & - & 0.955 & 0.831 & 0.915/0.913 & 0.647  \\
		CAC~\cite{CAC} & $\surd$ & $\times$ & 0.985 & 0.938 & 0.803 & 0.863/0.872 & 0.772  \\
		RPL~\cite{Reciprocal} & $\surd$ & $\times$ & 0.996 & 0.968 & 0.901 & 0.976/0.968 & 0.809  \\
		A-RPL-CS~\cite{A-RPL} & $\surd$ & $\times$ & 0.997 & 0.967 & 0.910 & 0.971/0.951 & 0.782  \\
		Hybrid~\cite{Hybrid} & $\surd$ & $\times$ & 0.995 & 0.947 & 0.950 & 0.962/0.955 & 0.793  \\
		PROSER~\cite{Placeholder} & $\surd$ & $\times$ & - & 0.943 & 0.891 & 0.960/0.953 & 0.693  \\
		EGT~\cite{extreme_geometric} & $\surd$ & $\times$ & - & 0.958 & 0.821 & 0.937/0.930 & 0.709  \\
		Capsule~\cite{Capsule} & $\surd$ & $\times$ & 0.992 & 0.956 & 0.835 & 0.888/0.889 & 0.715   \\
		S2OSC~$\dagger$~\cite{S2OSC} & $\times$ & $\surd$ & 0.995 & 0.936 & 0.855 & 0.910/0.809 & 0.714   \\
		\midrule
		A-RPL~\cite{A-RPL} & $\surd$ & $\times$ & 0.996 & 0.963 & 0.901 & 0.965/0.943 & 0.762  \\
		IT-OSR-ARPL & $\times$ & $\surd$ & \textbf{0.999} & 0.982 & 0.952 & 0.990/0.991 & 0.849   \\
		\midrule
		TransP & $\surd$ & $\times$ & 0.984 & 0.948 & 0.913 & 0.950/0.962 & 0.910   \\
		IT-OSR-TransP & $\times$ & $\surd$ & \textbf{0.999} & \textbf{0.983} & \textbf{0.965} & \textbf{0.991/0.993} & \textbf{0.943}   \\
		\bottomrule
	\end{tabular}
	\label{table:AUROC}
	\\[12pt]
	\centering
	\setlength{\abovecaptionskip}{0pt}
	\setlength{\belowcaptionskip}{10pt}
	\caption{Evaluation on closed-set classification (ACC) under the standard-dataset setting. The reported results are averaged over the same five trials as~\cite{OSRCI, GDFR, Hybrid, extreme_geometric}.}
	\begin{tabular}{m{2.6cm}<{\raggedright}m{1.2cm}<{\centering}m{1.2cm}<{\centering}m{1.5cm}<{\centering}m{1.5cm}<{\centering}m{1.8cm}<{\centering}m{2.2cm}<{\centering}m{2.2cm}<{\centering}}
		\toprule
		Method & Inductive & Transductive & MNIST & SVHN & CIFAR10 & CIFAR+10/+50 & TinyImageNet  \\
		\midrule
		SoftMax/
		OpenMax~\cite{openmax} & $\surd$ & $\times$ & 0.995 & 0.947 & 0.801 & - & -  \\
		OSRCI~\cite{OSRCI} & $\surd$ & $\times$ & 0.996 & 0.951 & 0.821 & - & -  \\
		CROSR~\cite{CROSR} & $\surd$ & $\times$ & 0.992 & 0.945 & 0.930 & - & -  \\
		C2AE~$\dagger$~\cite{C2AE} & $\surd$ & $\times$ & 0.992 & 0.936 & 0.910 & 0.919 & 0.430  \\
		CGDL~$\dagger$~\cite{CGDL} & $\surd$ & $\times$ & 0.996 & 0.942 & 0.912 & 0.914 & 0.445  \\
		GDFR~\cite{GDFR} & $\surd$ & $\times$ & - & 0.973 & 0.951 & 0.974 & 0.559  \\
		CAC~\cite{CAC} & $\surd$ & $\times$ & 0.998 & 0.970 & 0.934 & 0.952 & 0.759  \\
		RPL~$\dagger$~\cite{Reciprocal} & $\surd$ & $\times$ & 0.996 & 0.967 & 0.939 & 0.943 & 0.642  \\
		A-RPL-CS~$\dagger$~\cite{A-RPL} & $\surd$ & $\times$ & 0.997 & 0.971 & 0.953 & 0.956 & 0.678  \\
		Hybrid~$\dagger$~\cite{Hybrid} & $\surd$ & $\times$ & 0.995 & 0.962 & 0.926 & 0.937 & 0.612  \\
		PROSER~\cite{Placeholder} & $\surd$ & $\times$ & - & 0.964 & 0.926 & - & 0.521  \\
		EGT~\cite{extreme_geometric} & $\surd$ & $\times$ & - & 0.977 & 0.943 & 0.959 & 0.656  \\
		Capsule~$\dagger$~\cite{Capsule} & $\surd$ & $\times$ & 0.994 & \textbf{0.984} & 0.952 & 0.969 & 0.774   \\
		S2OSC~$\dagger$~\cite{S2OSC} & $\times$ & $\surd$ & 0.996 & 0.941 & 0.925 & 0.918 & 0.689   \\
		\midrule
		A-RPL~\cite{A-RPL} & $\surd$ & $\times$ & 0.996 & 0.971 & 0.952 & 0.955 & 0.679  \\
		IT-OSR-ARPL & $\times$ & $\surd$ & 0.996 & 0.972 & 0.953 & 0.955 & 0.785   \\
		\midrule
		TransP & $\surd$ & $\times$ & \textbf{0.997} & 0.980 & \textbf{0.988} & 0.987 & 0.937   \\
		IT-OSR-TransP & $\times$ & $\surd$ & \textbf{0.997} & 0.980 & \textbf{0.988} & \textbf{0.988} & \textbf{0.945}   \\
		\bottomrule
	\end{tabular}
	\label{table:ACC}
\end{table*}

\begin{table*}[t]
	\centering
	\setlength{\abovecaptionskip}{0pt}
	\setlength{\belowcaptionskip}{10pt}
	\caption{Evaluation on open-set classification (macro-F1 score) under the cross-dataset setting.}
	\begin{tabular}{m{2.6cm}<{\raggedright}m{1.4cm}<{\centering}m{1.4cm}<{\centering}m{2.5cm}<{\centering}m{2.5cm}<{\centering}m{2cm}<{\centering}m{2cm}<{\centering}}
		\toprule
		Dataset & Inductive & Transductive & ImageNet-crop & ImageNet-resize & LSUN-crop & LSUN-resize  \\
		\midrule
		SoftMax & $\surd$ & $\times$ & 0.639 & 0.653 & 0.642 & 0.647  \\
		OpenMax~\cite{openmax} & $\surd$ & $\times$ & 0.660 & 0.684 & 0.657 & 0.668  \\
		OSRCI~\cite{OSRCI} & $\surd$ & $\times$ & 0.636 & 0.635 & 0.650 & 0.648  \\
		CROSR~\cite{CROSR} & $\surd$ & $\times$ & 0.721 & 0.735 & 0.720 & 0.749  \\
		C2AE~~\cite{C2AE} & $\surd$ & $\times$ & 0.837 & 0.826 & 0.783 & 0.801   \\
		CGDL~\cite{CGDL} & $\surd$ & $\times$ & 0.840 & 0.832 & 0.806 & 0.812   \\
		GDFR~\cite{GDFR} & $\surd$ & $\times$ & 0.757 & 0.792 & 0.751 & 0.805   \\
		CAC~$\dagger$~\cite{CAC} & $\surd$ & $\times$ & 0.764 & 0.752 & 0.756 & 0.777   \\
		RPL~$\dagger$~\cite{Reciprocal} & $\surd$ & $\times$ & 0.811 & 0.810 & 0.846 & 0.820   \\
		A-RPL-CS~$\dagger$~\cite{A-RPL} & $\surd$ & $\times$ & 0.862 & 0.841 & 0.859 & 0.873   \\
		Hybrid~$\dagger$~\cite{Hybrid} & $\surd$ & $\times$ & 0.802 & 0.786 & 0.790 & 0.757   \\
		PROSER~\cite{Placeholder} & $\surd$ & $\times$ & 0.849 & 0.824 & 0.867 & 0.856   \\
		EGT~$\dagger$~\cite{extreme_geometric} & $\surd$ & $\times$ & 0.829 & 0.794 & 0.826 & 0.803   \\
		Capsule~\cite{Capsule} & $\surd$ & $\times$ & 0.857 & 0.834 & 0.868 & 0.882    \\
		S2OSC~$\dagger$~\cite{S2OSC} & $\times$ & $\surd$ & 0.828 & 0.810 & 0.832 & 0.806    \\
		\midrule
		A-RPL~$\dagger$~\cite{A-RPL} & $\surd$ & $\times$ & 0.858 & 0.830 & 0.845 & 0.867   \\
		IT-OSR-ARPL & $\times$ & $\surd$ & 0.939 & 0.910 & 0.921 & 0.908   \\
		\midrule
		TransP & $\surd$ & $\times$ & 0.881 & 0.870 & 0.908 & 0.891   \\
		IT-OSR-TransP & $\times$ & $\surd$ & \textbf{0.971} & \textbf{0.959} & \textbf{0.973} & \textbf{0.971}   \\
		\bottomrule
	\end{tabular}
	\label{table:F1}
\end{table*}

\subsection{Evaluation Under the Standard-Dataset Setting}
Aiming at both detecting unknown classes and classifying known classes, an effective method for open-set recognition should perform well on both the open-set detection task and the closed-set classification task~\cite{survey}. Thus we compare the two derived transductive OSR methods with 15 open-set recognition methods on both open-set detection and closed-set classification under the standard-dataset setting, and the results are reported in Tables~\ref{table:AUROC} and~\ref{table:ACC} respectively. The results marked with $\dagger$ are reproduced by their published codes or by ourselves because these results are unreported in their papers. 

As seen from Tables~\ref{table:AUROC} and~\ref{table:ACC}, compared with the two baseline inductive methods (A-RPL~\cite{A-RPL} and TransP), 
the performances of the two derived transductive OSR methods under the proposed IT-OSR framework are significantly improved on the relatively difficult datasets (\emph{e.g.} TinyImageNet), indicating that the proposed transductive OSR framework is able to boost the performances of inductive OSR methods. Besides, the two derived transductive methods perform better than other 15 existing OSR methods for both open-set detection and closed-set classification in most cases, demonstrating the effectiveness of the proposed framework for handling the open-set recognition task.

\subsection{Evaluation Under the Cross-Dataset Setting}
Comparing with the standard-dataset setting, there is a larger domain shift under the cross-dataset setting because the known classes and the unknown classes are from different datasets. 
We evaluate the two derived transductive OSR methods and the 15 existing OSR
methods respectively under the cross-dataset setting, and the corresponding results are reported in Table~\ref{table:F1}, where those marked with $\dagger$ are reproduced.

As seen from Table~\ref{table:F1}, both of the derived IT-OSR methods perform significantly better than their baseline inductive methods
as well as the other comparative methods. These results demonstrate that the proposed IT-OSR framework is insensitive to the relatively larger domain shift and it has a better generalization ability, probably because of the designed dual-space consistent sampling approach and the proposed conditional dual-adversarial generative network.

\subsection{Ablation Study}

\textbf{Ablation study on the proposed sampling approach and generative network.} To better analyze the effect of the explored dual-space consistent sampling approach (denoted DSCS) and the proposed conditional dual-adversarial generative network (denoted DGAN), we conduct an ablation study on DSCS and DGAN. The experiment is implemented in the IT-OSR-TransP method on the ImageNet-crop dataset under the cross-dataset setting, which compares the methods not only with or without (denoted w/o) DSCS and DGAN but also with the sampling approach in S2OSC~\cite{S2OSC} (denoted S2OSC-sampling) instead of DSCS, and the results are reported in Table~\ref{table:ablation}. Note that the methods w/o DSCS sample from the test set only by a scoring approach (\emph{i.e.} the Step S1 in the reliability sampling module), the methods w/o DGAN train a typical conditional GAN (\emph{i.e.} $G+F+D_1+C_1$) instead in the feature generation module. As seen from Table~\ref{table:ablation}, the comparison between IT-OSR-TransP w/o DGAN and IT-OSR-TransP demonstrates the effect of the proposed DGAN, and the comparisons of IT-OSR-TransP to both IT-OSR-TransP w/o DSCS and IT-OSR-TransP with S2OSC-sampling demonstrate that the explored DSCS is effective.

\begin{table}[t]
	\centering
	\setlength{\abovecaptionskip}{0pt}
	\setlength{\belowcaptionskip}{10pt}
	\caption{Comparison of macro-F1 score for ablation study on DSCS and DGAN}
	\begin{tabular}{m{5.6cm}<{\centering}m{2.4cm}<{\centering}}
		\toprule
		Method & macro-F1   \\
		\midrule
		TransP & 0.881      \\
		\midrule
		IT-OSR-TransP w/o DSCS nor DGAN  & 0.949  \\
		IT-OSR-TransP w/o DGAN & 0.964   \\
		IT-OSR-TransP w/o DSCS & 0.962   \\
		IT-OSR-TransP with S2OSC-sampling & 0.965   \\
		IT-OSR-TransP & 0.971   \\
		\bottomrule
	\end{tabular}
	\label{table:ablation}
\end{table}

\begin{table}[t]
	\centering
	\setlength{\abovecaptionskip}{0pt}
	\setlength{\belowcaptionskip}{10pt}
	\caption{Comparison of macro-F1 score for ablation study on $D_2$ and $C_2$ of the proposed DGAN}
	\begin{tabular}{m{5.6cm}<{\centering}m{2.4cm}<{\centering}}
		\toprule
		Architecture & macro-F1   \\
		\midrule
		$G+F+D_1+C_1$  & 0.964  \\
		$G+F+D_1+C_1+C_2$ & 0.966   \\
		$G+F+D_1+D_2+C_1$ & 0.969   \\
		IT-OSR-TransP & 0.971   \\
		\bottomrule
	\end{tabular}
	\label{table:ablation2}
\end{table}

\textbf{Ablation study on the architecture of the proposed generative network.} Besides, to better analyze the effect of each network part in the proposed DGAN, we also conduct an ablation study on $D_2$ and $C_2$ ($G$, $F$, $D_1$ and $C_1$ are indispensible), which is also implemented in the IT-OSR-TransP method on the ImageNet-crop dataset under the cross-dataset setting. The results are reported in Table~\ref{table:ablation2}. As seen from Table~\ref{table:ablation2}, the comparison between $G+F+D_1+C_1+C_2$ and IT-OSR-TransP demonstrates that the discriminator $D_2$ in the proposed DGAN is important for improving the discrimination of the known/unknown-class features, and the comparison between $G+F+D_1+D_2+C_1$ and IT-OSR-TransP demonstrates that the classifier $C_2$ in the proposed DGAN is also effective for better performance.

\section{Conclusion}   \label{Conclusion}
In this paper, we propose the Iterative Transductive OSR framework, IT-OSR, which iteratively performs three explored modules: the reliability sampling module, the feature generation module, and the baseline update module. We explore the dual-space consistent sampling approach for selecting a relatively reliable subset from the pseudo-labeled test samples in the reliability sampling module, and the conditional dual-adversarial generative network for balancing the number of the pseudo-labeled unknown-class test samples and that of the known-class samples in the feature generation module of the explored IT-OSR. Any inductive OSR method can be seamlessly embedded into IT-OSR for alleviating the domain shift problem. We further derive two novel transductive OSR methods under the explored IT-OSR framework, and extensive experimental results demonstrate the effectiveness of IT-OSR.

%
\newpage

\bibliographystyle{IEEEtran}
\bibliography{egbib_20220117}

\newpage

\begin{IEEEbiographynophoto}{Jiayin Sun} is currently pursuing the ph.D. degree in pattern recognition and intelligence systems with the National Laboratory of Pattern Recognition, Inastitute of Automation, Chinese Academy of Sciences. Her current research interests include machine learning and its applications, particularly in the open-set recognition.
\end{IEEEbiographynophoto}
\vspace{-200 mm}
\begin{IEEEbiographynophoto}{Quilei Dong} received the B.S. degree in automation from Northeastern University, Shenyang, China, in 2003 and the Ph.D. degree from the Institute of Automation, Chinese Academy of Sciences, Beijing, China, in 2008. He is currently a Professor with the National Laboratory of Pattern Recognition, Institute of Automation, Chinese Academy of Sciences. His current research interests include 3-D computer vision and pattern classification.
\end{IEEEbiographynophoto}

\end{document}